\documentclass[letterpaper, 10 pt, conference]{ieeeconf}  

\IEEEoverridecommandlockouts                              

\usepackage{amsmath} 
\usepackage{amssymb}  
\usepackage[ruled,vlined]{algorithm2e}
\usepackage{booktabs}
\usepackage{graphicx}
\usepackage{listings}
\usepackage{hyperref}
\title{\LARGE \bf
Annotation-Free One-Shot Imitation Learning\\ for Multi-Step Manipulation Tasks
}

\author{Vijja Wichitwechkarn$^{1*}$, Emlyn Williams$^{2}$, Charles Fox$^{2}$, Ruchi Choudhary$^{1}$
\thanks{$^{1}$Dept. of Engineering, University of Cambridge, Cambridge CB2 1PZ}%
\thanks{$^{2}$School of Computer Science, University of Lincoln, Lincoln LN6 7TS}%
\thanks{$^{*}$Corresponding Author: {\tt\small vw273@cam.ac.uk}}%
}

\begin{document}

\maketitle
\thispagestyle{empty}
\pagestyle{empty}


\begin{abstract}
Recent advances in one-shot imitation learning have enabled robots to acquire new manipulation skills from a single human demonstration. While existing methods achieve strong performance on single-step tasks, they remain limited in their ability to handle long-horizon, multi-step tasks without additional model training or manual annotation. We propose a method that can be applied to this setting provided a single demonstration without additional model training or manual annotation. We evaluated our method on multi-step and single-step manipulation tasks where our method achieves an average success rate of 82.5\% and 90\%, respectively. Our method matches and exceeds the performance of the baselines in both these cases. We also compare the performance and computational efficiency of alternative pre-trained feature extractors within our framework. Project repository: \href{https://github.com/vijja-w/AF-OSIL-MS}{https://github.com/vijja-w/AF-OSIL-MS}

\end{abstract}

\section{INTRODUCTION}

Recent advances in imitation learning have enabled robots to perform increasingly complex tasks. However, these methods still require hundreds to thousands of demonstrations per task \cite{zhao2023learning, chi2023diffusion, zhao2024aloha, goyal2023rvt}, making them impractical for real-world deployment. In contrast, humans can quickly acquire and deploy new skills. To bring robots closer to this level of deployability, it is crucial to minimize the effort required to acquire new manipulation skills. Ideally, robots should be able to learn and deploy new skills from a single demonstration, without the need for additional data collection, model training, or manual annotation.

One-shot imitation learning (OSIL) offers this possibility by teaching a robot a new behavior from a single demonstration, with a large group of methods based on pose estimation combined with visual servoing and trajectory replay. These include methods that train models for segmentation and visual servoing \cite{valassakis2022demonstrate}, and methods that train pose estimation models \cite{vitiello2023one}. Canonical models of objects can also be trained to allow generalization to different objects within the same category \cite{biza2023one}. OSIL can also be achieved without training models, using classical visual features with pre-trained matchers \cite{wang2025one}, or correspondence algorithms with pre-trained features \cite{di2024dinobot}. However, these methods typically assume the availability of object masks \cite{vitiello2023one, wang2025one, biza2023one} and therefore require manual keypoint annotation for pre-trained segmentation models such as SAM2 \cite{ravi2024sam}. Additionally, to extend replay to multistep tasks, demonstrations must be manually decomposed into multiple single-step demonstrations, which are then further annotated by selecting the bottleneck poses and segmenting the trajectories for replay.

There are also other approaches to OSIL beyond trajectory replay. For example, trajectory optimization based on keypoints obtained through pre-trained models and heuristics has been used for OSIL\cite{tang2025functo}. World models can also be trained to predict latent trajectories that are decoded into physical waypoints for OSIL \cite{goswami2025osvi}. These methods however both require manual decomposition to be applied to multi-step tasks similar to trajectory replay based methods \cite{tang2025functo, goswami2025osvi}. A human-to-robot alignment method has also been demonstrated for OSIL, requiring a vision encoder and action policy model to be trained \cite{kedia2025one}. Dynamical systems can also be applied to OSIL \cite{li2025elastic}, but this imposes strong priors such as attractors and smoothness constraints that limit their ability to capture topologically complex behaviors. An invariance-based method has also been shown to natively support arbitrary trajectories and multi-step tasks without additional annotation. This method however requires training a model that is used in an optimization-based framework to generate end-effector poses \cite{zhang2024one}.

There are also few shot methods ($\sim\!10$ demonstrations) that use large pre-trained models and do not require model training or manual annotations \cite{di2024keypoint}. Variations improve upon this by relying on manual language annotations to provide additional context \cite{team2025gemini}. There are also methods that achieve few shot learning without manual annotation, but require models to be trained \cite{vosylius2024instant, fang2024keypoint}.

We propose a trajectory-replay-based imitation learning approach that requires only a single demonstration and does not depend on additional data collection, model training, or manual annotation. We demonstrate the effectiveness of our method on the manipulation of everyday objects, comparing against baselines on both multi-step and single-step tasks. In addition, we evaluate the performance of different pre-trained vision encoders within our framework and also evaluate their computational requirements in terms of runtime and memory usage. The source code can be found here: \href{https://github.com/vijja-w/AF-OSIL-MS}{https://github.com/vijja-w/AF-OSIL-MS}.

\section{METHODS}
\subsection{\textbf{Problem Formulation}}

We consider the following formulation for one-shot imitation learning. We focus on a single robot arm with a wrist camera. A demonstration trajectory is defined as

\begin{equation}
\tau \;=\; \{(x_t, T_t, g_t)\}_{t=1}^{N},
\end{equation}

\noindent
where

\begin{equation}
T_t \in SE(3),
\end{equation}

\noindent
with $x_t$ denoting the camera image observation, $T_t$ the end-effector pose, $g_t$ the gripper state, and $N$ the total number of timesteps in the demonstration trajectory.

\noindent
We decompose the demonstration trajectory $\tau$ into $K$ ordered subtasks, each of which is a contiguous subsequence of the trajectory, using a decomposition function $f_{\text{decompose}}$:
\begin{equation}
(\tau^{(1)}, \tau^{(2)}, \dots, \tau^{(K)}) \;=\; f_{\text{decompose}}(\tau).
\end{equation}

\noindent
The function $f_{\text{decompose}}$ selects $K$ ordered, non-overlapping contiguous segments of the trajectory, possibly omitting intervals where no relevant action is taken, such as pauses. Each subtask $\tau^{(k)}$ is defined as a contiguous subsequence
\begin{equation}
\tau^{(k)} \;=\; \{(x_t, T_t, g_t)\}_{t=s_k}^{e_k},
\end{equation}

\noindent
where the start and end indices satisfy
\begin{equation}
1 \leq s_1 < e_1 \le s_2 < e_2 \le \cdots \le s_K < e_K \leq N.
\end{equation}

\noindent
Thus, the subtasks preserve the temporal order of the original trajectory, though they need not cover every timestep. For each subtask $k$, we select a \emph{keyframe} index $t_k^\star \in \{s_k,\dots,e_k\}$ using a keyframe selector function $f_{\text{keyframe}}$:
\begin{equation}
t_k^\star \;=\; f_{\text{keyframe}}\!\left(\tau^{(k)}\right),
\end{equation}
and denote the corresponding state as
\begin{equation}
x_k^\star := x_{t_k^\star}, \qquad T_k^\star := T_{t_k^\star}, \qquad g_k^\star := g_{t_k^\star}.
\end{equation}

\noindent
Each subtask $k$ is defined by two consecutive phases: the alignment phase and the execution phase.

In the \textbf{alignment phase}, the robot is aligned to the selected keyframe by minimizing the discrepancy between the current camera image $x$ and the keyframe image $x_k^\star$. This involves estimating a relative camera transform
\begin{equation}
\Delta T^{\text{cam}} = f_{\text{align}}(x, x_k^\star),
\end{equation}

\noindent
where $f_{\text{align}}$ is a visual alignment procedure. The relative motion in camera coordinates is converted into the end-effector frame as
\begin{equation}
\Delta T^{\text{ee}} =
T_{\text{cam},\text{ee}} \cdot
\Delta T^{\text{cam}} \cdot
T_{\text{cam},\text{ee}}^{-1},
\end{equation}

\noindent
where the hand-eye calibration is given by a fixed transform $T_{\text{cam},\text{ee}} \in SE(3)$, denoting the camera pose expressed in the end-effector frame. The end-effector pose is updated as
\begin{equation}
T \;\leftarrow\; T \cdot \Delta T^{\text{ee}},
\end{equation}
which is iterated until the alignment error between $x$ and $x_k^\star$ falls below a set threshold.

This is followed by the \textbf{execution phase} where the robot replays the remaining demonstration actions of subtask $k$. Each action consists of a gripper command $g_j$ and a relative motion $\Delta T_j$ derived from the demonstration trajectory:
\begin{equation}
\Delta T_j = T_j^{-1} T_{j+1}, \qquad j = t_k^\star, \dots, e_k - 1.
\end{equation}
This yields the control sequence,
\begin{equation}
\mathbf{u}^{(k)} =
\big(a_{t_k^\star}, a_{t_k^\star+1}, \dots, a_{e_k-1}\big),
\qquad a_j = (g_j, \Delta T_j).
\end{equation}

\noindent
These relative actions are applied sequentially starting from the pose obtained after the alignment phase:
\begin{equation}
T \;\leftarrow\; T \cdot \Delta T_j, 
\qquad g \;\leftarrow\; g_j.
\end{equation}

\begin{figure}[h] 
    \centering
    \includegraphics[width=1.0\columnwidth]{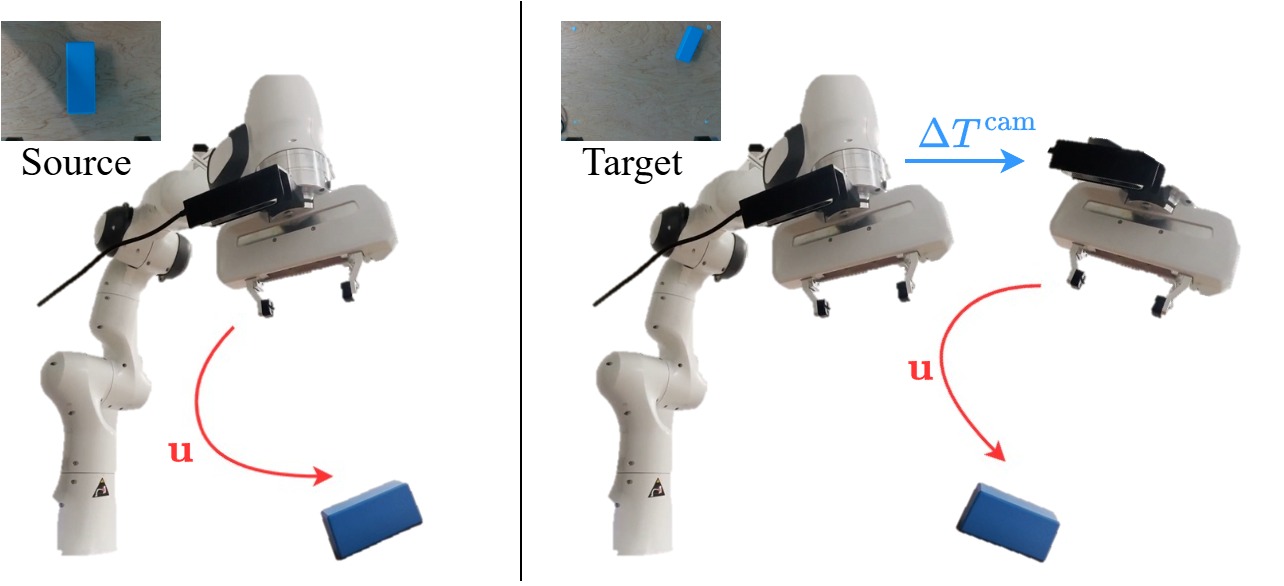} 
    \caption{Diagram of alignment and trajectory replay. (\textbf{Left}) A demonstration provides a keyframe image (source) and its associated control sequence $\mathbf{u}$. (\textbf{Right}) At inference, the current observation (target) is used to estimate the camera pose change $\Delta T^{\text{cam}}$ needed to align the camera with the visual target in the same way as the source image. The control sequence $\mathbf{u}$ is then executed relative to this aligned pose.}

    \label{fig:align_execute}
\end{figure}

\noindent
Thus, each subtask is an atomic unit consisting of a closed-loop alignment to its keyframe followed by a replay of the corresponding action sequence. This is illustrated in Figure~\ref{fig:align_execute}. Although our formulation specifies control in terms of position waypoints, the same approach could also be realized with velocity commands. We choose waypoints because they are easy to interpret and compatible with standard controllers that accept pose targets.

\noindent
At inference, the robot executes the demonstration by iterating through the sequence of subtasks. The procedure is summarized in Algorithm~\ref{alg:execution}.

\begin{algorithm}[t]
\DontPrintSemicolon
\KwIn{For each subtask $k=1,\dots,K$: keyframe image $x_k^\star$ and open-loop control sequence $\mathbf{u}^{(k)}$}
\KwOut{Executed trajectory}
Initialize robot state $(x, T, g)$\;
\For{$k \gets 1$ \KwTo $K$}{
    Align$(x, x_k^\star)$\;
    Execute$(\mathbf{u}^{(k)})$\;
}
\caption{One-shot imitation execution}
\label{alg:execution}
\end{algorithm}

\subsection{\textbf{Our Method}}
\begin{figure*}[h] 
    \centering
    \includegraphics[width=1.0\linewidth]{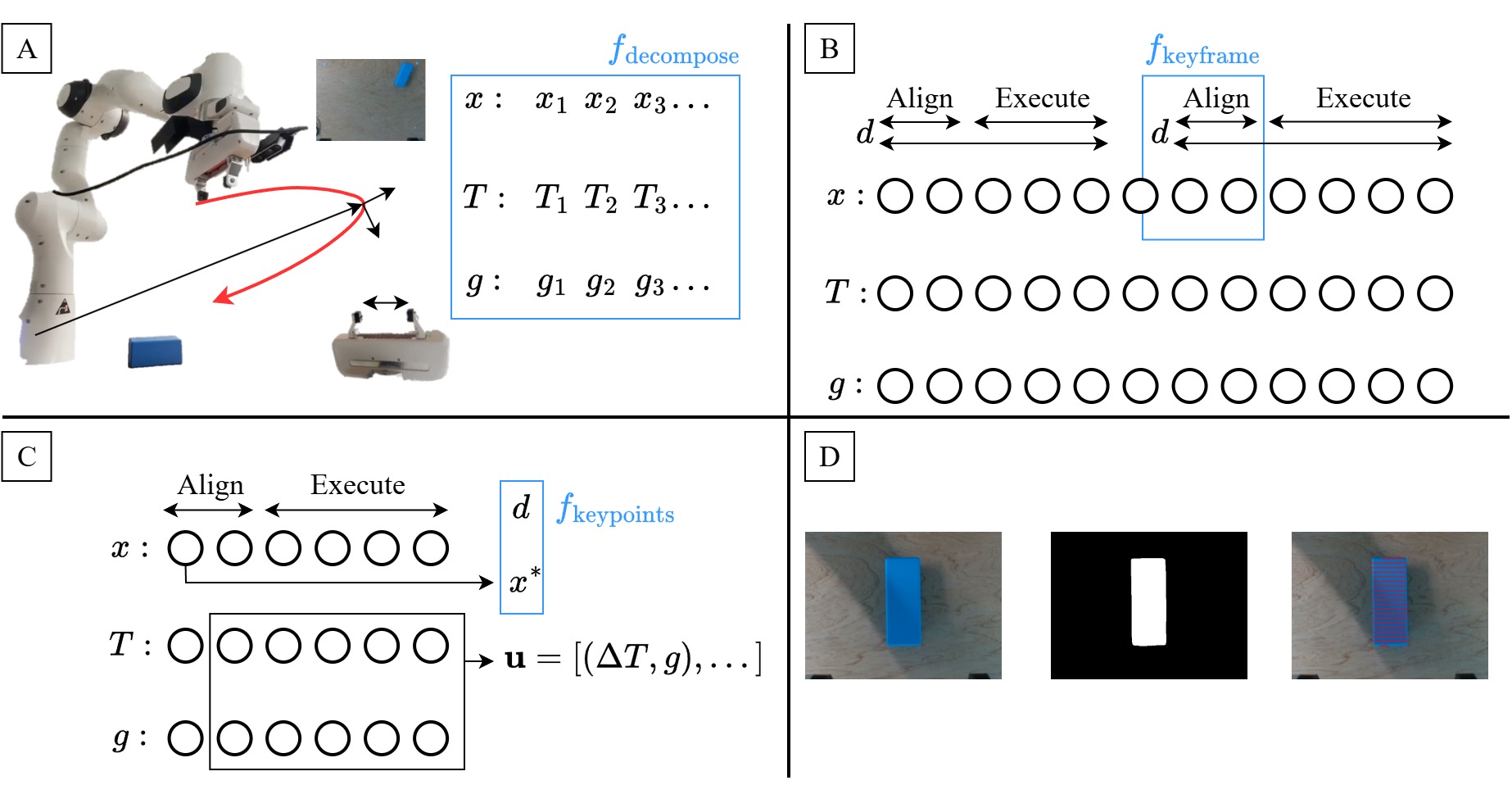} 
    \caption{Overview of our method. (\textbf{A}) A demonstration consists of a sequence of wrist-camera images $x$, end-effector poses $T$, and gripper widths $g$. As indicated by the blue box, these are provided as context to $f_{\text{decompose}}$. (\textbf{B}) $f_{\text{decompose}}$ outputs subtask partitions, each described by a visual target $d$ and time-stamps marking the alignment and execution phases. As shown in the blue box, the alignment-phase image sequence together with $d$ are used as context for $f_{\text{keyframe}}$, applied separately to each subtask. (\textbf{C}) $f_{\text{keyframe}}$ outputs a keyframe $x^{}$ selected from the alignment phase. The control sequence $\mathbf{u}$ is obtained by computing relative poses $\Delta T$ from the remainder of the alignment phase through to the end of the execution phase. As shown in the blue box, $x^{}$ and $d$ are then passed as context to $f_{\text{keypoints}}$. This is also applied separately for each subtask. (\textbf{D}) $f_{\text{keypoints}}$ outputs a mask of the visual target described by $d$, which is uniformly sampled to produce the keypoints shown as red dots.}
    \label{fig:method}
\end{figure*}

Given this formulation and a single demonstration $\tau$, our task is to define the decomposition function $f_{\text{decompose}}$ for obtaining the atomic subtasks from $\tau$, the keyframe selector $f_{\text{keyframe}}$ that determines the bottleneck keyframe and consequently the control sequence, and the alignment function $f_{\text{align}}$ for aligning the camera at inference. The aim of our work is to do this in a manner that requires no additional model training or manual annotation. We use a single robot arm with a parallel jaw gripper and an RGB-D wrist camera and specify $g_{t}$ using the gripper width. An overview of our method is shown in Figure~\ref{fig:method}.

\subsubsection{\textbf{Decomposition $f_{\text{decompose}}$}}
We implement subtask decomposition using a pre-trained vision-language model (VLM), particularly Gemini~2.5 Pro \cite{comanici2025gemini}.  The VLM input  consists of (i) the demonstration RGB video $X$ downsampled to $1$~fps, (ii) the gripper context $G = \{g_{t}, v_{t}, |v_{t}|\}_{t=1}^{N}$ containing the gripper width, velocity $v_{t}$ and speed $|v_{t}|$ (derived from the demonstration and downsampled to $1$~Hz), and (iii) a natural language prompt $p_{\text{decompose}}$. Formally,
\begin{equation}
\big\{(\hat{s}_k^{\text{(A)}}, \hat{e}_k^{\text{(A)}}, \hat{s}_k^{\text{(E)}}, \hat{e}_k^{\text{(E)}}, d_k)\big\}_{k=1}^K
= \mathrm{VLM}(p_{\text{decompose}}, X, G),
\end{equation}
\noindent
where $(\hat{s}_k^{\text{(A)}}, \hat{e}_k^{\text{(A)}})$ and $(\hat{s}_k^{\text{(E)}}, \hat{e}_k^{\text{(E)}})$ are the predicted start and end indices of the alignment and execution phases of subtask $k$, respectively, and $d_k$ is a description of the visual target in the alignment phase. These outputs are then used to recover the formal decomposition,
\begin{equation}
(\tau^{(1)}, \tau^{(2)}, \dots, \tau^{(K)}) \;=\; f_{\text{decompose}}(\tau),
\end{equation}
\noindent
where each $\tau^{(k)}$ corresponds to the subsequence of $\tau$ delimited by $(\hat{s}_k^{\text{(A)}}, \hat{e}_k^{\text{(E)}})$. The associated description $d_k$ is later used by $f_{\text{keyframe}}$ to select a bottleneck keyframe within the identified alignment phase for each subtask. The prompt $p_{\text{decompose}}$ is provided in Appendix~\ref{appendix:prompt_decompose}.

\subsubsection{\textbf{Keyframe selection $f_{\text{keyframe}}$}}
Given the phase boundaries and visual target description $d_k$ from $f_{\text{decompose}}$, we crop the original demonstration RGB video $X$ to the alignment interval of each subtask $k$, denoted $X_{k}^{(A)} = \{x_t\}_{t=\hat{s}_k^{\text{(A)}}}^{\hat{e}_k^{\text{(A)}}}$. This cropped segment is downsampled to $5$~fps and passed to the VLM together with $d_k$ and a natural language prompt $p_{\text{keyframe}}$. Formally,
\begin{equation}
t_k^\star = \mathrm{VLM}(p_{\text{keyframe}}, X_{k}^{(A)}, d_k),
\end{equation}
\noindent
where $t_k^\star$ is the index of the selected keyframe within $\tau^{(k)}$ that best corresponds to the visual target described in $d_{k}$. The corresponding keyframe observation is then
\begin{equation}
x_k^\star := x_{t_k^\star}.
\end{equation}
\noindent
This completes the definition of $f_{\text{keyframe}}$:
\begin{equation}
t_k^\star = f_{\text{keyframe}}\!\left(\tau^{(k)}, d_k\right).
\end{equation}
\noindent
The prompt $p_{\text{keyframe}}$ is provided in Appendix~\ref{appendix:prompt_keyframe}.

\subsubsection{\textbf{Alignment $f_{\text{align}}$}}
For each subtask $k$, the alignment function $f_{\text{align}}$ takes as input the keyframe image $x_k^\star$ and the current camera observation $x$. It outputs a relative camera motion $\Delta T^{\text{cam}}$ that, when applied, aligns the current view with the visual target configuration depicted in $x_k^\star$. We implement $f_{\text{align}}$ using the following steps: \\

\paragraph{\textbf{Source Keypoint Extraction}}
We provide $x_k^\star$ together with the description $d_k$ and a prompt $p_{\text{keypoints}}$ to Gemini~2.5 Flash. The model directly outputs a segmentation mask for the visual target described in $d_k$, which is uniformly sampled and projected to 3D keypoints $P_k^\star = \{\mathbf{p}_i^\star \in \mathbb{R}^3\}$ using the known camera intrinsics. The prompt $p_{\text{keypoints}}$ is provided in Appendix~\ref{appendix:prompt_keypoints}.

\paragraph{\textbf{Target Keypoint Extraction}}
At inference time, given the current camera image $x$, we compute dense visual features using a pre-trained vision model (we used the Spatial Perception Encoder \cite{bolya2025perception} which was the state of the art at the time experimental work was conducted).  We find the best correspondences between the features from the sampled source and target keypoints, yielding 2D correspondences in the target image $x$. The centroid of these correspondences is passed to SAM2~\cite{ravi2024sam} (sam2.1\_hiera\_large), which segments the corresponding object in $x$. Uniform sampling from the mask and back-projection using intrinsics yields a set of target 3D keypoints $P = \{\mathbf{p}_i \in \mathbb{R}^3\}$. We used the centroid for simplicity but other methods can be utilized such as the means of a Gaussian mixture model.


\paragraph{\textbf{Pose Estimation}}
We estimate relative camera motion $\Delta T^{\text{cam}}$ by aligning $P_k^\star$ and $P$ via generalized ICP \cite{segal2009generalized}:
\begin{equation}
\Delta T^{\text{cam}} = \mathrm{gICP}(P_k^\star, P).
\end{equation}
To improve robustness, we apply RANSAC with this process. We terminate the alignment phase when the Frobenius norm between $\Delta T^{\text{cam}}$ and the identity matrix falls below a threshold. The alignment function is thus,
\begin{equation}
\Delta T^{\text{cam}} = f_{\text{align}}(x, x_k^\star, d_k).
\end{equation}

\subsection{\textbf{Baseline Methods}}
Our proposed method is a trajectory replay method designed for OSIL on multi-step tasks without any additional model training or manual annotation. This distinguishes it from existing approaches that do not have this property. We therefore compare against two types of baseline methods: (i) trajectory replay methods that require only one demonstration and no further training but are limited to single-step tasks and require manual annotation, and (ii) few-shot methods that leverage multiple demonstrations without additional training and are capable of handling multi-step tasks.

\subsubsection{\textbf{Single-step Tasks}}

\paragraph{\textbf{DinoBot} \cite{di2024dinobot}}
Given a demonstration with a single subtask, the keyframe is manually selected. Semantic keypoints between this source image and a target image are automatically extracted using best-buddies nearest neighbors applied to DINOv2 \cite{oquab2023dinov2} features. Least-squares optimization is then used to compute $\Delta T^{\text{cam}}$.

\paragraph{\textbf{ODIL} \cite{wang2025one}}
We adapt the method from ODIL (One-shot Dual-arm Imitation Learning), using the part that concerns single-arm alignment using a wrist camera. Given a demonstration with a single subtask, the keyframe is manually selected along with keypoints that are used by SAM2 \cite{ravi2024sam} to obtain the source image mask. For feature extraction, ODIL employs GPU-optimized SIFT descriptors with a pre-trained matcher (LightGlue \cite{lindenberger2023lightglue}). Unidirectional filtering is applied to retain only matches that lie within the source mask. Pose estimation is then performed using a weighted Kabsch algorithm using the matching weights from LightGlue. This is combined with RANSAC.\\

\subsubsection{\textbf{Multi-Step Tasks}}
\begin{table}[h]
\centering
\caption{Success rates (\%) on multi-step tasks over 10 trials. The best results are shown in bold.}
\begin{tabular*}{\columnwidth}{@{\extracolsep{\fill}}lcccc}
\toprule
\textbf{Method} & \textbf{Stack Blocks} & \textbf{Sort Fruits} & \textbf{Tea Prep.} & \textbf{Wipe} \\
\midrule
KAT     & 0   & 0   & 0   & 0   \\
GEMINI  & 0   & 0   & 0   & 0   \\
Ours    & $\mathbf{90}$  & $\mathbf{80}$  & $\mathbf{90}$  & $\mathbf{70}$  \\
\bottomrule
\end{tabular*}
\label{tab:multi_step_rate}
\end{table}

\begin{table}[h]
\centering
\caption{Number of iterations our method required for convergence during the alignment phase, averaged over 10 trials for each visual target in the multi-step tasks.}

\begin{tabular*}{\columnwidth}{@{\extracolsep{\fill}}cccc}
\toprule
 \textbf{Block} & \textbf{Cup} & \textbf{Coaster} & \textbf{Tea Bag} \\
\midrule
  $7.8 \pm 3.1$ &  $6.9 \pm 2.4$ & $6.3 \pm 4.6$  & $6.9 \pm 2.4$  \\

\midrule
\addlinespace[1.5ex]
\midrule
\textbf{Eraser} & \textbf{Smudge} & \textbf{Strawberry} & \textbf{Bucket} \\
\midrule
 $5.1 \pm 1.2$  & $22.4 \pm 12.2$  & $8.1 \pm 4.0$  &  $6.1 \pm 1.8$ \\

\bottomrule

\end{tabular*}
\label{tab:multi_step_n}
\end{table}

\begin{figure}[h] 
    \centering
    \includegraphics[width=1.0\linewidth]{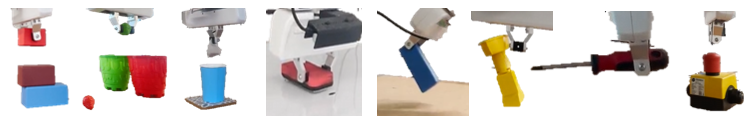} 
    \caption{Representative images of the tasks used in our experiments. From left to right: block stacking, fruit sorting, tea preparation, smudge erasing, block flipping, tower knocking, screwdriver pickup, and button pressing.
    }
    \label{fig:tasks_all}
\end{figure}

\paragraph{\textbf{KAT} \cite{di2024keypoint}} (Keypoint Action Tokens) leverage multiple demonstrations to condition a VLM to perform manipulation tasks, which can be multi-step. We use the same VLM used in our method as opposed to GPT4-Turbo that was used in the original work. Given $N_{d}$ demonstrations, 3D semantic keypoints are automatically extracted using the same method as DinoBot. These are treated as the `input' for each demonstration. To obtain the corresponding `output', each demonstration is temporally subsampled to $N_{a}$ timesteps. At each timestep, the robot state is represented by the gripper width together with the end-effector pose. Rather than encoding the pose $T$ directly, KAT represents the end-effector using the positions of three fixed reference points defined relative to the tool center point. These trajectories are treated as the `output' for each demonstration. At test time, the current observation image is converted into semantic 3D keypoints which are used as the `current input'. These input-output pairs, along with the test time current input are used as context for a VLM that is used to predict the waypoints to execute. In our experiments, we follow the settings recommended by the original work, using $N_{d}=10$ demonstrations and $N_{a}=20$. For further detail, we refer the reader to the original work \cite{di2024keypoint}.

\paragraph{\textbf{GEMINI} \cite{team2025gemini}} follows the same methodology as KAT, but with two key differences: (i) the end-effector pose is represented using quaternions rather than reference points, and (ii) additional natural language annotations describing the task are provided as part of the VLM input. Since the official prompt specification and experimental setting is unavailable, we adopt the same ones used in KAT for our comparison.

\section{RESULTS}
We evaluated our approach on a Franka Emika Panda robotic arm equipped with an Intel RealSense D435 RGB-D camera, and using a workstation with an NVidia GeForce RTX 4070 SUPER GPU and an Intel\textregistered Core\texttrademark  i7-14700KF. We conducted three types of experiments: (i) multi-step tasks comparing our method against multi-step task baselines, (ii) single-step tasks comparing against single-step task trajectory-replay baselines, and (iii) component comparisons analyzing different pre-trained models used for correspondence. Representative images of the tasks used in our experiments are shown in Figure \ref{fig:tasks_all}.

\subsection{\textbf{Multi-step tasks}}
\label{section:multi_step_tasks}
\noindent
We evaluated our method against the KAT and GEMINI baselines on four multistep manipulation tasks, each is repeated for 10 trials per task. The success rates are shown in Table~\ref{tab:multi_step_rate}. The number of iterations required for convergence during the alignment phase is shown in Table~\ref{tab:multi_step_n}. These are averaged over 10 trials for each subtask. The multi-step tasks used are described as follows.

\begin{itemize}
    \item \textbf{Block stacking:} sequentially stack three blocks. This is a pick-and-place task with rigid objects with a well-defined geometric shape. This task has four subtasks.
    \item \textbf{Fruit sorting:} pick and place a green toy strawberry into a green bucket, and a red toy strawberry into a red bucket. This requires handling rigid but irregularly shaped objects. This task has four subtasks.
    \item \textbf{Tea preparation:} place a tea cup on a coaster, and insert a tea bag into the cup. This involves a deformable and irregularly shaped object (the tea bag). This task has four subtasks.
    \item \textbf{Whiteboard Wiping:} pick up a whiteboard eraser and wipe a smudge on a whiteboard. This is a contact-rich manipulation task, with variability since each smudge is different. This task has two subtasks.
\end{itemize}

\subsection{\textbf{Single-step tasks}}

\begin{table}[h]
\centering
\caption{Success rates (\%) on single-step tasks over 10 trials. Best results are shown in bold.}
\begin{tabular*}{\columnwidth}{@{\extracolsep{\fill}}lcccc}
\toprule
\textbf{Method} & \textbf{Flip} & \textbf{Knock} & \textbf{Pick} & \textbf{Press} \\
\midrule
DinoBot &  10 & 40  &  60 &  \textbf{70} \\
ODIL    &  0 &  40 &  0 &  0 \\
Ours    & \textbf{100}  &  \textbf{100} & \textbf{90}  & \textbf{70}  \\
\bottomrule
\end{tabular*}
\label{tab:single_step_rate}
\end{table}

\begin{table}[h!]
\centering
\caption{Iterations required for convergence during the alignment phase, averaged over 10 trials for single-step tasks.}

\begin{tabular*}{\columnwidth}{@{\extracolsep{\fill}}lcccc}
\toprule
\textbf{Method} & \textbf{Flip} & \textbf{Knock} & \textbf{Pick} & \textbf{Press} \\
\midrule
DinoBot     &  $6.0 \pm 0.0$ &  $7.0 \pm 3.5$  &  $3.8 \pm 0.7$  & $8.9 \pm 5.4$   \\
ODIL  &  - &  $4.0 \pm 1.0$  &  -  &   - \\
Ours    &  $6.6 \pm 2.2$  & $6.0 \pm 1.7$  & $6.0 \pm 1.8$  & $20.3 \pm 10.7$  \\

\bottomrule

\end{tabular*}
\label{tab:single_step_n}
\end{table}

\noindent
We evaluated our method against DinoBot and ODIL on four single-step manipulation tasks, each is repeated for 10 trials per task. These include flipping a block, knocking a tower over, picking up a screw driver and pressing a button. The success rates are shown in Table~\ref{tab:single_step_rate}. The number of iterations required for convergence during the alignment phase is shown in Table~\ref{tab:single_step_n}. These are averaged over 10 trials if there are failures only the successful trials are used.

\subsection{Component Comparisons}

We compare variations of our method that differ in the choice of pre-trained models used for correspondence.  Specifically, we evaluate Aspanformer \cite{chen2022aspanformer}, which automatically identifies correspondences by suggesting keypoints and matching them. We also evaluate DINOv2 \cite{oquab2023dinov2} and CleanDIFT \cite{stracke2025cleandift}, which provide alternative pre-trained features to our method which uses the Perception Encoder (spatial) \cite{bolya2025perception}. For context, we also include the baseline methods DinoBot and ODIL in our comparison.

\begin{figure*}[h] 
    \centering
    \includegraphics[width=1.0\linewidth]{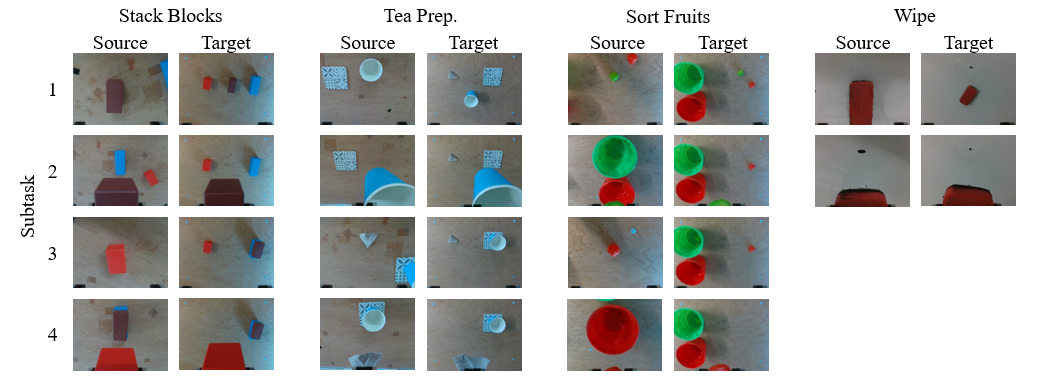} 
    \caption{The source image (demonstration keyframe) and target image (test time observation) for each subtask of the multi-step tasks.
    }
    \label{fig:source_target}
\end{figure*}

To ensure conditions similar to deployment, we evaluate these methods using source and target images from the multi-step tasks described in Section~\ref{section:multi_step_tasks}. These images are shown in Figure \ref{fig:source_target}. The pose estimation accuracy of these variations is evaluated in terms of translation and rotation error, relative to the known ground truth obtained from iterating the alignment phase until convergence. These results are shown in Table~\ref{tab:pose_estimation_error}. Table~\ref{tab:compute} shows the peak VRAM usage and average runtime over 10 runs.

\begin{table*}[h]
\centering
\caption{Translation error $E_{t}$ (m) and rotation error $E_{R}$ (degrees) for each method given the source and target images for subtasks from the multi-step tasks. The best results are shown in bold and failures are denoted using dashes.}
\begin{tabular*}{\textwidth}{@{\extracolsep{\fill}}lcccccccccccc}
\toprule
\textbf{Subtask} 
& \multicolumn{2}{c}{\textbf{Aspanformer}} 
& \multicolumn{2}{c}{\textbf{DinoV2}} 
& \multicolumn{2}{c}{\textbf{CleanDIFT}} 
& \multicolumn{2}{c}{\textbf{Ours}} 
& \multicolumn{2}{c}{\textbf{DinoBot}} 
& \multicolumn{2}{c}{\textbf{ODIL}} \\
\cmidrule(lr){2-3}
\cmidrule(lr){4-5}
\cmidrule(lr){6-7}
\cmidrule(lr){8-9}
\cmidrule(lr){10-11}
\cmidrule(lr){12-13}
& $E_t$ & $E_R$ 
& $E_t$ & $E_R$ 
& $E_t$ & $E_R$ 
& $E_t$ & $E_R$ 
& $E_t$ & $E_R$ 
& $E_t$ & $E_R$ \\
\midrule
Stack Blocks 1    & 0.0661 & 15.88 & \textbf{0.0277} & \textbf{8.87}  & 0.0655 & 15.60 & \textbf{0.0277} & 8.99  & 0.0750 & 18.22 & -     & - \\
Stack Blocks 2     & -      & -     & 0.1127 & 11.04 & \textbf{0.1061} & \textbf{4.72}  & \textbf{0.1061} & \textbf{4.72}  & 0.4070 & 56.15 & -     & - \\
Stack Blocks 3      & 0.0854 & \textbf{12.46} & \textbf{0.0693} & 34.18 & 0.0848 & 12.91 & 0.0853 & 12.49 & 0.3247 & 25.23 & -     & - \\
Stack Blocks 4 & \textbf{0.1081} & 15.66 & 0.3371 & 83.93 & 0.2836 & 28.39 & 0.1655 & \textbf{6.65}  & 0.1736 & 19.49 & -     & - \\
\midrule
Tea Prep. 1            & -      & -     & 0.2204 & 59.03 & 0.2205 & 59.06 & \textbf{0.1264} & \textbf{31.62} & 0.3109 & 64.44 & -     & - \\
Tea Prep. 2        & 0.1185 & \textbf{6.09}  & 0.1185 & 6.10  & 0.1185 & \textbf{6.09}  & 0.1185 & \textbf{6.09}  & 0.1813 & 44.49 & \textbf{0.1176} & 6.26 \\
Tea Prep. 3         & -      & -     & 0.2560 & 41.79 & 0.2343 & 37.66 & \textbf{0.0395} & \textbf{18.83} & 0.1675 & 44.67 & -     & - \\
Tea Prep. 4    & 0.0402 & 5.85  & 0.0402 & 5.89  & 0.0402 & \textbf{5.83}  & 0.0402 & 5.85  & -      & -     & \textbf{0.0211} & 38.0 \\
\midrule

Sort Fruits 1    & 0.0387 & 8.38  & \textbf{0.0384} & \textbf{8.34}  & 0.0386 & 8.40  & 0.0385 & \textbf{8.34}  & 0.0891  & 16.75  & 0.0777 & 20.88 \\
Sort Fruits 2    & 0.0896 & \textbf{1.11}  & \textbf{0.0895} & \textbf{1.11}  & \textbf{0.0895} & \textbf{1.11}  & 0.0896 & 1.13  & 0.2032  & 27.24  & 0.1018 & 3.19 \\
Sort Fruits 3    & - & - & 0.1437 & 52.81  & 0.1438 & 52.85  & 0.1438 & 52.83  & \textbf{0.1298}  & \textbf{17.59}  & - & - \\
Sort Fruits 4    & 0.1111 & 1.87  & 0.2007 & 9.98  & 0.1111 & 1.87  & 0.1111 & \textbf{1.86}  & \textbf{0.0868}  & 17.09  & 0.1196 & 3.16 \\

\midrule
Wipe 1    & - & -  & 0.1855 & \textbf{7.34}  & \textbf{0.0453} & 20.91  & \textbf{0.0453} & 20.93  & -  & -  & 0.1606 & 47.29 \\
Wipe 2    & - & -  & 0.1279 & 62.22  & \textbf{0.0458} & 67.53  & 0.0682 & 77.86  & 0.1053  & \textbf{55.03}  & - & - \\

\bottomrule
\end{tabular*}
\label{tab:pose_estimation_error}
\end{table*}

\begin{table}[h]
\centering
\caption{The time it takes and the peak VRAM required to compute $\Delta T^{\text{cam}}$ for variations of our method and baseline methods.}

\begin{tabular*}{\columnwidth}{@{\extracolsep{\fill}}lcc}
\toprule
\textbf{Method} & \textbf{Time / s} & \textbf{VRAM / GB}  \\
\midrule
Aspanformer & $ 0.460 \pm 0.047$ & 3.75  \\
DinoV2 & $0.362 \pm 0.004$ &  4.20\\
CleanDIFT & $0.530 \pm 0.042$  &  12.16\\
Ours & $ 0.351 \pm 0.004$ &   7.22\\

\midrule
DinoBot &  $1.694 \pm 0.025$ & 1.39 \\
ODIL    &  $0.207 \pm 0.006$ & 0.91\\

\bottomrule

\end{tabular*}
\label{tab:compute}
\end{table}

\section{DISCUSSION}

On multi-step tasks, our method achieves an average success rate of 82.5\%, in contrast to the baselines which both achieved 0\%. The baselines generally failed because they were not sufficiently precise to reproduce the demonstrated interactions with objects. For example, in the wiping task the baseline could execute a wiping motion but failed to first pick up the eraser; on other tasks it often selected the wrong object as the target and also failed to grasp it. 
This may be due to the use of the Gemini model in place of GPT-4~Turbo as in the original implementation.

Despite the strong overall performance of our method, we observed several characteristic failure modes. Some errors arose from shadows in the RGB-D sensor images, which occasionally produced poor point-cloud reconstructions for the cup in the tea preparation task. Failure also occurs when the predicted $\Delta T^{\text{cam}}$ overshot, pushing the visual target out of the camera’s field of view. In addition, imperfect convergence due to our thresholding sometimes led to suboptimal alignment. This can be alleviated by reducing the threshold at the cost of increased number of iterations for convergence. Instead of executing the full predicted $\Delta T^{\text{cam}}$ in one step, an incremental strategy that moves in the correct direction while recomputing $\Delta T^{\text{cam}}$ online would mitigate overshooting and reduce alignment time. Additionally, since the gripper control sequence is not adaptive, execution errors cannot be corrected once they occur. For example, when grasping an irregularly shaped strawberry, the object could slip out of the gripper as it closed. Since such failures are not detected during execution, the system continues blindly, causing the grasp to fail without recovery.

The number of iterations required for convergence was generally manageable (5-8 iterations), but was higher for the smudge erasing task, likely due to variations in smudge shape. This issue was not present for the tea bag where shape variation exists, but to a much smaller degree.

On single-step tasks, our method also performs well, achieving an average success rate of 90\% and consistently matching or exceeding baselines that require manual annotation. The ODIL baseline failed frequently because there were insufficient keypoints detected within the source mask, with SIFT often being confused by background textures. DinoBot performed better since its DINO features provide stronger semantic cues, but still failed in cases of poor alignment or when irrelevant keypoints were selected.

Compared to these baselines, our method typically required a higher number of iterations to converge. This is likely due to noise in point-cloud sampling and shadow artifacts from the RGB-D camera, which introduced variations between runs. When ODIL or DinoBot succeeded, they generally converged in fewer iterations, likely benefiting from more semantically meaningful keypoints. The button task in particular exhibited higher mean iteration counts and greater variance, which is likely due to shadowing effects around the button region.

Our method compares favorably against variations of our framework with different pre-trained feature extractors, and the DinoBot and ODIL baselines. Methods relying on keypoint proposals, such as Aspanformer, ODIL, and DinoBot, exhibited failure cases where insufficient correspondences were detected.  In contrast, our method and its DinoV2 and CleanDIFT variations always succeeded, due to uniform sampling to obtain keypoints. 

Among these, our method typically achieves the lowest translation and rotation error or when this is not the case its error is very close to that of DinoV2 and CleanDIFT variations. In terms of computational requirements, our method achieved the lowest run-time among the variations of our framework. 
Since memory usage was not a limiting factor, this made it the most efficient variation overall. DinoBot and ODIL had the lowest memory footprint, with ODIL running very quickly while DinoBot was comparatively slow, likely due to the provided implementation.

\section{Conclusion}
We introduced a one-shot imitation learning framework that enables robots to perform long-horizon multi-step tasks from a single demonstration without requiring additional data collection, model training, or manual annotation. Our method outperforms or matches the performance of few-shot baselines on multi-step tasks and single-shot baselines that require manual annotations on single-step tasks.

While our approach demonstrates strong performance, it also has important limitations. The method is not designed to generalize to unseen objects, and performance is restricted to objects with similar point-cloud structure. For the same reasons, it will also struggle under occlusion. Future work in incorporating 3D reconstruction methods may address this and may also improve pose estimation. Our current approach relies on uniform mask sampling, which can introduce noise. Incorporating semantic keypoints provides an interesting direction for improvement.

Although our method supports a wide variety of tasks, not all manipulations are supported. For example, tasks that require relative motions between objects such as pushing one object closer to another. Since execution is open-loop there is no error correction once the execution phase begins. The method also cannot be applied to cases where identical instances of the same object are present in the scene or when the visual targets are not in view. Integrating VLM modules to check task completion, re-route execution, engage `scanning'/`searching' behaviour or provide further reasoning to address these limitations would be an interesting direction for future work. 

\section*{APPENDIX}

\subsection{Prompts}
\label{appendix:prompts}
The following are the full prompts that are used in our method.

\subsubsection{\textbf{Subtask Decomposition $p_{\text{decompose}}$}}
\label{appendix:prompt_decompose}
\noindent

\textit{Context:
This is the point of view of a wrist camera of a robot gripper. The robot itself is not in view. The fingers of the gripper are at the bottom of the frame, seen as two black pads. The robot gripper starts with nothing in it.}

\noindent
\textit{Sensor readings:
\{gripper\_states\}}

\noindent
\textit{This behaviour is generated by calling the function:}

\begin{verbatim}
do_task(visual_target, waypoints):
   align_gripper_to(visual_target)
   execute_stored_waypoints(waypoints)
   pullback()
\end{verbatim}

\textit{align\_gripper\_to(visual\_target):
will have the robot align the camera and hence the gripper with the provided 'visual\_target'}

\textit{execute\_stored\_waypoints(waypoints):
Will make the robot blindly execute the stored waypoints relative to its current pose. This is done blindly so
there can be no relative movements towards another visual\_target, that is only handled in the align\_gripper\_to phase.}

\textit{pullback():
will cause the robot to blindly disengage and prepare itself for the next task by moving back.
This is done blindly so there can be no relative movements towards another visual\_target, that is only handled in the 
align\_gripper\_to phase.}

\textit{visual\_target = `string visually describing the object appearance'
waypoints = [ [ gripper\_pose\_6d, gripper\_width ], [ gripper\_pose\_6d, gripper\_width ], ... ]}

\textit{Important constraints that must all be satisfied: }

\textit{The do\_task can only do one primitive task in one call: eg. pick, place, press, etc. It can not do multiple tasks in one call, eg. pick and place, these must be two separate do\_task calls.
The execute\_stored\_waypoints phase must not include relative movements to a different\ visual\_target than the one in the align\_to\_gripper step before it.
The pullback phase must not include relative movements to a different visual\_target than the one in the align\_to\_gripper step before it.
The timestamp ranges must not overlap
The timestamp must be within the time range of the video and sensor readings.
The pullback phase must not be confused with the align\_gripper\_to phase of the next task. The boundary should be close to the pullback event (more time allocated to the next align\_gripper\_to phase). This is because the pullback phase is a short disengage while the align\_gripper\_to phase of the next task should be a longer movement relative to the next visual target to align it. }

\textit{Your objective:
Decompose the robot's behaviour into this do\_task call that must satisfy all of the above important constraints. What is the robot doing in this video? How many times is this function called? What is the visual\_target for each call (description of the appearance)? If you use colour in your description only use one word for the colour. You are allowed to use words like light and dark in addition to the one word for the colour. What is the timestamp for each phase of the call: align\_gripper, execute\_stored\_waypoints, pullback?}

\textit{For each call tell me the reasoning, include all of the following: }

\textit{Refer to sensor readings
Discuss honestly the task and whether this can potentially be decomposed further, remember things like pick and place must be decomposed into two separate do\_task calls. 
Describe honestly the motion in the execute\_stored\_waypoints phase and whether this is relative to a different visual\_target and hence is not executed blindly.
Describe honestly the motion in the pullback phase and whether this is relative to a different visual\_target and hence is not executed blindly.
Why this visual target, name the other potential visual targets seen and you selected this one.
Discuss the decision for the boundary between the pullback phase and the next align\_gripper\_to phase (if any).}

\textit{It is fine if you think you have made a mistake, be honest and flag this. There will be a chance to correct this later. Give timestamps range in the format start then finish `MM:SS-MM:SS'}

\subsubsection{\textbf{Keyframe Selection $p_{\text{keyframe}}$}}
\label{appendix:prompt_keyframe}

\noindent
\textit{Context:
This is the point of view of a wrist camera of a robot gripper.
The robot itself is not in view.
The fingers of the gripper are at the bottom of the frame, seen as two black pads.
The robot gripper may potentially be gripping and holding something.
Ignore all information about the object the gripper is gripping (if any) focus only on the visual target.}

\textit{Visual Target: \{source\_object\_description\}}

\textit{Select the timestamp where the entire silhouette (outer contour) of the visual target is fully visible within the frame,
with no part cropped by the image boundaries or occluded by the silhouette of other objects, for instance the object
held by the gripper, if any. Make sure there is some margin between the silhouette and the image boundaries or other objects.   }  

\textit{Give timestamps in the format `MM:SS'}

\subsubsection{\textbf{Segmentation $p_{\text{keypoints}}$}}
\label{appendix:prompt_keypoints}

\noindent
\textit{Give the segmentation masks for the \{source\_object\_description\}}




\bibliographystyle{IEEEtran}   
\bibliography{refs}

\end{document}